# Road detection via a dual-task network based on cross-layer graph fusion modules

Zican Hu, Wurui Shi, Hongkun Liu, and Xueyun Chen

*Abstract*—Road detection based on remote sensing images is of great significance to intelligent traffic management. The performances of the mainstream road detection methods are mainly determined by their extracted features, whose richness and robustness can be enhanced by fusing features of different types and cross-layer connections. However, the features in the existing mainstream model frameworks are often similar in the same layer by the single-task training, and the traditional cross-layer fusion ways are too simple to obtain an efficient effect, so more complex fusion ways besides concatenation and addition deserve to be explored. Aiming at the above defects, we propose a dual-task network (DTnet) for road detection and cross-layer graph fusion module (CGM): the DTnet consists of two parallel branches for road area and edge detection, respectively, while enhancing the feature diversity by fusing features between two branches through our designed feature bridge modules (FBM). The CGM improves the cross-layer fusion effect by a complex feature stream graph, and four graph patterns are evaluated. Experimental results on three public datasets demonstrate that our method effectively improves the final detection result. The used code is available at https://github.com/huzican695/DTnet.

## I. INTRODUCTION

Remote sensing road detection is an essential and challenging task in semantic segmentation, widely applied in urban planning, road monitoring, and other fields. In recent years, methods based on neural networks have become the mainstream, and the variety and robustness of extracted features play a critical role in model performances. The existing feature optimization methods include depth and width increment [1], self-attentions, cross-layer features fusion, etc. However, it remains a challenging task to satisfy practical needs.

Mainstream target detection methods have proposed many methods to enhance the network feature diversity and robustness performance for improving accuracy. Szegedy *et al.*[1] increased the depth and width of the network to learn more various features, respectively. Chen *et al.* [2] used a set of multi-scale dilated convolutions to obtain more target spatial texture features. Zhao *et al.* [3] enriched global contextual features through a pyramid pooling module of different scales. Lin *et al.* [4] introduced a chained residual pooling module for getting more background features.

The traditional cross-layer fusion between encoder and decoder was first proposed by Ronneberger *et al*. [5], which is beneficial for recovering spatial contextual information in the deep layers of the network, widely used in target detection methods [6][7]. Zao *et al.* [8] combined the features at the same level in the encoder network to fuse with the decoder features, which preserved more detailed information.

However, in the above single-task networks, the features of the same layer are often found to be too similar to obtain richer. Naturally, the thought of designing a dual-task network may be reasonable and beneficial for feature diversity. A series of researches have demonstrated that adding auxiliary tasks to the model has facilitated performance improvement [9]-[12]. Yao *et al.* [9] introduced a multi-stage framework for extracting the road surface and road centerline and combined the results of both stages to improve road segmentation performance. Lu *et al.* [10] added a centerline extraction decoder and Li *et al.*[11] added a task of matching the semantic constraint angle matrix of images, both of which make the extracted roads smoother and more complete. Liu *et al.* [12] first proposed a model to achieve all three tasks simultaneously: road surface segmentation, road edge detection, and road centerline extraction, which promote each other. However, none of the above approaches explored the effects of fusion between different task features. Besides, in the cross-layer fused processing, features of encoder layers contain some low-level spatial information, such as corners and edges. In contrast, features in decoder layers contain more discriminative information for the segmentation, i.e., more representative semantic information [13]. Introducing these encoder features into decoder features like traditional concatenation or addition operations is so simple that it may bring background noises, influence the performance and robustness of the feature. It leaves a research gap to explore more complex fusion modules between cross-layer for improvement.

Aiming at the above defects, in this letter, we propose a dual-task network (DTnet) based on feature bridge modules (FBM) and cross-layer graph fusion modules (CGM): DTnet contains two parallel branches, both are connected by our designed FBMs, while they execute the area and edge detection respectively. CGMs are introduced in the backbone of the main branch in order to promote cross-layer features fusion.

The dual-task network structure enhances the feature diversity and robustness of the same layers, and the feature performances of the fused layers are improved by the CGMs

This work was supported by the National Natural Science Foundation of China under Grant 62061002. (Corresponding author: Xueyun Chen.)

The authors are with the School of Electrical Engineering, Guangxi University, Nanning 530004, China (e-mail: 20140043@gxu.edu.cn)

and FBMs. We experiment on three public remote sensing datasets, and the results show superior performances of DTnet and positive effects of FBMs and CGMs.

The main contributions of this letter are as follows:

(1) Propose a novel dual-task network architecture, enhancing feature diversity and robustness by fusing the features from the branches assigned to different tasks.

(2) Present a new cross-layer graph fusion module, which improves the fusing performance of the cross-layer features via a complex feature stream graph. Four different graph fusion patterns are evaluated.

(3) Propose a novel feature bridge module, which builds a feature fusing bridge between two branches to promote features fused.

The remainder of this letter is as follows. Section II introduces our method in detail, including the DTnet, CGMs, and FBMs. Section III shows the experimental results and related analysis, and Section IV concludes.

## II. OUR APPROACH

### A. Network Structure

The dual-task network (DTnet) is divided into two coherent branches, as shown in Fig. 1: the main branch (E+D) and the side branch (E1+D1), whose tasks are different from each other, which can guide the network to generate different types of features and enhance the diversity and robustness of the features. The side branch task is road edge detection, so it can provide rich edge information for the main branch and help in edge recovery for road detection. Both E and E1 consist of four down-sampling residual blocks, D consists of one up-sampling residual block and one residual block alternately four times, and D1 consists of four up-sampling residual blocks. The down-sampling residual block output in E fuses with the up-sampling residual block output in D by the cross-layer graph fusion modules (CGM) to recover spatial contextual information in the deep layers features. The down-sampling residual block output in E and the residual block output in D are fused with the down-sampling residual block output in E1 and the up-sampling residual block output in D1 at the same level, respectively, by the feature bridge module (FBM).

### B. Cross-layer Graph Fusion Module

We designed the cross-layer graph fusion modules for cross-layer feature fused processing, as shown in Fig. 2. The idea of the module is to add spatial information to the decoder (deep) features $D^i$ and reinforce the semantic information of the encoder (shallow) features $E^i$ for bridging the semantic gap.

There are two strategies to add spatial information in $D^i$, as shown in A and B of Fig. 2. For A, the module uses $E^i$ and $D^i$ to generate spatial information features by multiple convolutions. For B, we generate a non-salient region map of $D^i$ by the *(1-x)* operation and then multiply the map with $E^i$ to extract the spatial information features of the corresponding regions in $E^i$. Finally, we add the spatial information features to $D^i$ to complement the spatial information needed. There are also two ways to reinforce semantic information on $E^i$, as shown in C and D of Fig. 2. For C, we use $D^i$ to directly multiply $E^i$ to reinforce semantic information of $E^i$ channel by channel. For D, we obtain a semantic reinforcement mask by performing the function $P(x): R^{H \times W \times C} \to R^{H \times W \times 1}$ on $D^i$ and then use the mask to reinforce the semantic information in the $E^i$. (A, C), (A, D), (B, C), (B, D) are combined to form the four structures (a), (b), (c), and (d) of CGM.

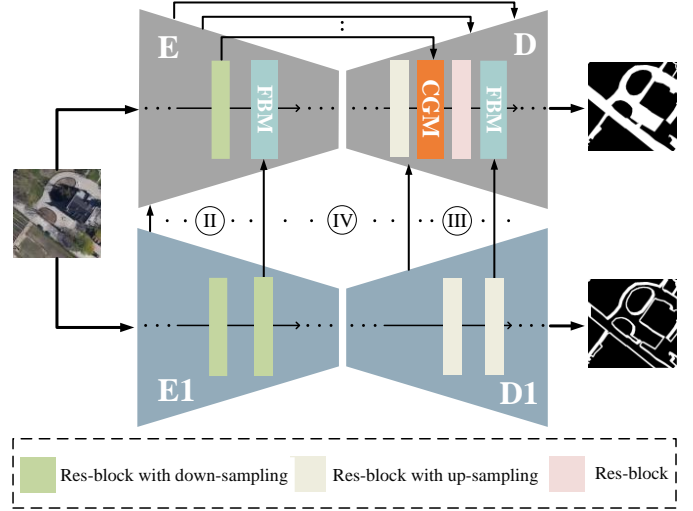

**Fig. 1.** Dual-task network (DTnet) structure. E and D are the encoder and decoder of the main branch, respectively. E1 and D1 are the encoder and decoder of the side branch. The features of E and D are fused by the cross-layer graph fusion module (CGM), and the features between two branches are fused by the feature fusion module (FBM). I, II, III, and IV are different fused placements, denoting all layers (II+III), encoder layers, decoder layers, and middle layers of the network.

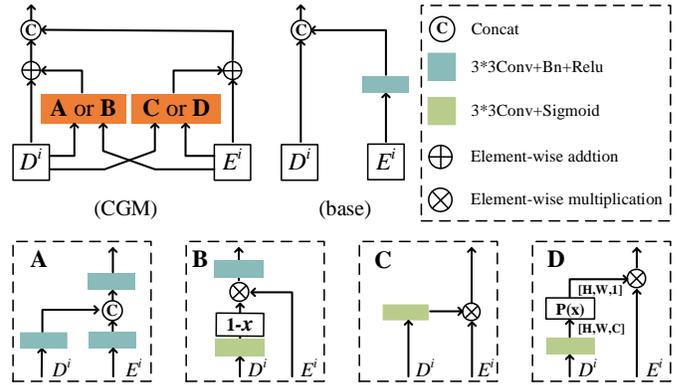

**Fig. 2.** The cross-layer graph fusion module (CGM). $E^i$ and $D^i$ are the encoder and decoder features, respectively, with the same size. (base) is the traditional cross-layer fusion method. $P(x)$ operation is expressed by Eq. (1).

$$D(x) = mean(x, dim = C), \ P(x) = \frac{D(x)}{max(D(x), dim = [H, W])} \quad (1)$$

Where feature $x \in R^{H \times W \times C}$. $P(x)$ is a simple function for compressed channels and normalization.

### C. Feature Bridge Module

We designed two types of feature bridge modules for the feature fusion process of two branches, as shown in Fig. 3 (c)

and (d), to promote features fused between the two branches of DTnet and enrich the main branch feature information by side branch. Class (c) can be used in any position between the two branches in the DTnet. We first map the side branch features ($EI^i$ or $DI^i$) into a mask by $P(x)$ operations and then use the mask multiply with the main branch features ($E^i$ or $D^i$) to enhance its edge information. Finally, concatenate the result with the original features ($E^i$ or $D^i$) and output. Class (d) is mainly used to fuse features (deep features) between D and D1. Deep features lack spatial information, so reinforcement by directly using side branch features may not be effective. Therefore, in class (d), we first add necessary spatial information to deep features $D^i$ through concatenating shallow features $E^i$, but may introduce noise information. Then we obtain a mask by performing the $Q(x)$ operation on the side branch features $DI^i$ and multiply it with the deep features $D^i$ to enhance the edge information and filter out the noise information. Finally, we concatenate the result with the original features $D^i$ and output.

$$Q(x) = softmax(mean(x^2, dim = C)) \qquad (2)$$

where feature $x \in R^{H \times W \times C}$. $Q(x)$ has the same function as $P(x)$: compressed channels and normalization, but it contains square and softmax operations, which allows features to be sparse and suppresses redundant information.

### D. Hybrid Loss Function

The loss function used in our method is computed via a linear combination of the three-tiered objective functions.

$$L_{sum}(x,y) = \frac{1}{N}\sum_{i=1}^{N}(a_1 L_{ce,i} + a_2 L_{iou,i} + a_3 L_{fl,i}) \qquad (3)$$

$$L_{ce,i} = -t_i log p_i - (1 - t_i) log(1 - p_i) \qquad (4)$$

$$L_{iou,i} = \frac{t_i p_i}{t_i + p_i - t_i p_i + C} \qquad (5)$$

$$L_{fl,i} = -\lambda(1-p_i)^\gamma t_i' log(p_i') - (1-\lambda)p_i'^\gamma (1-t_i') log(1-p_i') \qquad (6)$$

where, $L_{ce,i}$, $L_{iou,i}$, and $L_{fl,i}$ denote the cross-entropy loss, IOU loss [14], and Focal loss [15], respectively. $t_i$, $t_i'$, $p_i$, and $p_i'$ denote the road area label, road edge label, and prediction of the main branch and side branch, respectively. $N$ is the batch size, $C$ is the microscopic constants to avoid computing overflow, $\alpha_{1,2,3}$ is the empirically determined weight parameter, set to 1. $\lambda$ in $L_{fl,i}$ is used to balance the importance of positive/negative examples, set to 0.75, and $\gamma$ is a hyperparameter, set to 2.

The $L_{iou,i}$ focuses the network more on the target's area segmentation. $L_{fl,i}$ is the loss function for the side branch network, which can alleviate the significant problem of class imbalance in the edge segmentation.

## III. EXPERIMENTAL RESULTS AND ANALYSIS

### A. Dataset and Image Preprocessing

The method outlined in this letter has been verified on the Munich road dataset [16], the Massachusetts road dataset [17], and the LoveDA dataset [18]. The Munich road dataset contains

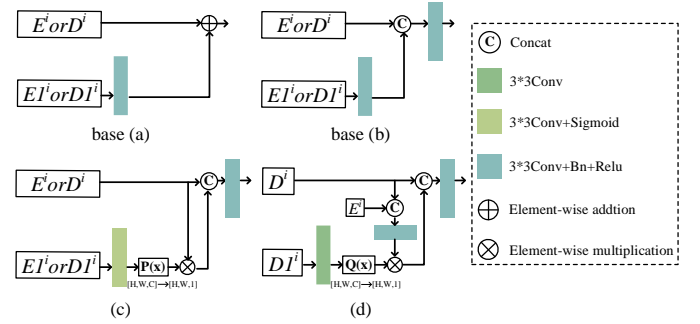

**Fig. 3.** Feature bridge module (FBM). $E^i$ and $D^i$ are the features of the encoder and decoder in the main branch, $EI^i$ and $DI^i$ are that of the side branch. base(a) and base(b) are the traditional feature fusion modules, (c) and (d) are the FBMs proposed in this letter. Module base(a), base(b), and (c) can be used in any position of feature fusion between the two branches, while module (d) is only used in the decoder location feature fusion of two branches. $Q(x)$ operation is expressed by Eq. (2).

20 remotely sensed images with a resolution of $5616 \times 3744$, but it lacks road calibration. Therefore, we carry out the calibration work on the road area. The Massachusetts road dataset contains 1171 remotely sensed images with a resolution of $1500 \times 1500$ and corresponding road calibration maps. LoveDA dataset contains 5987 remotely sensed images at 1024 x 1024 resolution. We randomly crop the above three datasets because the original image's data in the dataset is too large to be directly trained. In the Munich dataset, we obtained 533 images with a resolution of $512 \times 512$ and reshaped them to a resolution of $256 \times 256$ by bilinear interpolation, the training set is 484 images, and the test set is 49 images. In the Massachusetts road dataset, we obtained 6000 images with a resolution of $256 \times 256$, 5400 images as the training set and 600 images as the test set. In the LoveDA dataset, we chose the Nanjing city dataset and took the road objects as the labeled objects. Because its test set had no labels, we used the validation set to test and eliminate the samples that did not contain roads, then we reshaped them to a resolution of $512 \times 512$ by bilinear interpolation, with 976 images in the training set and 294 images in the test set.

Four metrics are used to evaluate the detection performance of the network, including IOU, Precision, Recall, and F1–score:

$$IOU = \frac{TP}{TP+FP+FN} \qquad (7)$$

$$Precision = \frac{TP}{TP+FP} \qquad (8)$$

$$Recall = \frac{TP}{TP+FN} \qquad (9)$$

$$F1 = 2 * \frac{precision*recall}{precision+recall} \qquad (10)$$

where TP is true positive, TF is true negative, FP is false positive, and FN is false negative.

### B. Experimental Results

In order to verify the effectiveness of our methods, a series of comparison experiments were conducted on the three public databases. Ablation experiments are carried out on Munich road

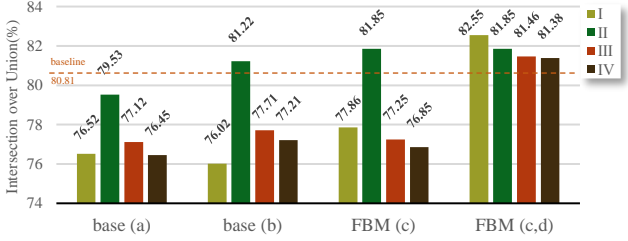

**Fig. 4.** Road detection results for using different feature bridge modules (FBMs) in the different positions of DTnet. base (a), and (b) denote traditional feature fusion modules. FBM (c) is class (c) FBM, and I, II, III, and IV are different fused positions between two branches of DTnet, denoting all layers, encoder, decoder, and middle layers. FBM (c, d) indicates that the features of the encoder (E and E1) and decoder (D and D1) layers in the fusion layers between two branches are fused using FBM (c) and (d), respectively.

TABLE I
ROADS DETECTION RESULTS OF DIFFERENT METHODS WITH AND WITHOUT SIDE BRANCH

| Methods | Munich (%) | | | |
|---|---|---|---|---|
| | IOU | F1 | Recall | Precision |
| ESPnet v2[20] | 65.36 | 77.51 | 76.72 | 79.71 |
| ESPnet v2 + Side_B | **72.43** | **83.07** | **85.17** | **82.34** |
| PSPnet[3] | 75.45 | 84.85 | 82.59 | 88.50 |
| PSPnet + Side_B | **77.42** | **86.40** | **84.40** | **89.76** |
| Unet[5] | 79.90 | 87.99 | 87.02 | 90.20 |
| Unet + Side_B | **81.82** | **89.35** | **88.59** | **91.32** |
| DeepLab v3+ | 80.84 | 88.63 | 86.35 | 91.98 |
| DeepLab v3+ + Side_B | **82.22** | **89.60** | **87.65** | **92.98** |

*Note: + Side_B means adding a side branch whose task is edge detection

TABLE II
ROADS DETECTION RESULTS OF THE NETWORK (E+D) USING DIFFERENT CROSS-LAYER FUSION MODULE

| Cross-layer fusing method | Munich (%) | | | |
|---|---|---|---|---|
| | IOU | F1 | Recall | Precision |
| baseline: CGM(base) | 80.81 | 88.05 | 85.81 | 91.55 |
| CGM(a) | **82.53** | **90.56** | 87.67 | **93.56** |
| CGM(b) | 81.99 | 90.11 | **87.83** | 92.51 |
| CGM(c) | 82.12 | 90.02 | 87.14 | 93.11 |
| CGM(d) | 82.04 | 90.16 | 87.44 | 93.06 |

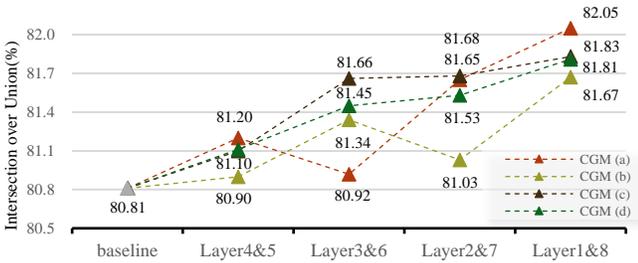

**Fig. 5.** Road detection results for different cross-layers of the network (E+D) use different CGMs. Layer#&# denotes using CGM for fusion between layer# and layer#.

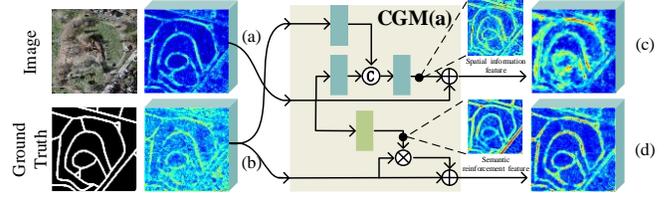

**Fig. 6.** Feature heat maps, (a): the features of D. (b): the features of E. (c) and (e): output of (a) and (b) after processing by CGM.

dataset to evaluate the contribution of the dual-task network (DTnet), cross-layer graph fusion modules (CGM), and feature bridge modules (FBM) to the model. The structure of baseline is encoder (E) + decoder (D) + CGM (base) in all experiments.

We set up experiments to find the most effective FBM and its best position between the two branches of DTnet. The results are shown in Fig. 4. The best is that the encoders (E and E1) and decoders (D and D1) between two branches use FBM (c) and (d), respectively, for feature fusion, and IOU is improved by 1.74% compared to baseline. Inappropriate feature bridge modules may add much redundant information to the feature and lower detection accuracy. Our designed FBMs in this letter use a mask to reinforce the original edge information of the features so that no additional redundant information is added.

Based on experiment FBM (c, d)(I), we applied our proposed DTnet framework to several single-task networks. The results are shown in Table I. IOU increased by 7.07%, 1.97%, 1.92%, and 1.38%, respectively, thus indicating that our method can enhance feature diversity for performance improvement.

We conduct experiments with different cross-layer fused methods on the network (E+D) to demonstrate the effectiveness of CGMs. Table II lists the experimental results, and it shows that the network using our designed CGMs improved IOU by 1.72%, 1.14%, 1.31%, and 1.23% compared to the baseline. Thus, CGM can effectively improve the feature performance of the cross-layer fusion. Visualization in Fig. 6 (c) and (d) indicate that after CGM(a) processing, the spatial information in the original deep features (a) is significantly increased, and the representation ability of the original shallow features (b) is enhanced. Fusing features as (c) and (d) in Fig. 6, which have a similar semantic level, is more effective than fusing (a) and (b).

Besides, we used CGMs in turn for fusion between different cross-layer features. As the results show in Fig5, the trend in IOU for different CGMs is up in general. Hence the larger the span of fused layers, the better the network performance improvement effect by CGMs, which also proves that the simple fusion method between long cross-layer features limits the improvement of network performance and our CGM can improve this problem.

Ultimately, DTnet uses class (a) CGM for cross-layer feature fusion while using class (c) FBM and class (d) FBM for feature fusion of encoders and decoders between two branches, respectively.

We compared our proposed with some mainstream semantic segmentation methods on the three datasets, and Table III shows the results. It can be seen that our model is significantly better than the other models. Moreover, we visualize some results in Fig. 7, and our proposed method is more accurate in segmenting difficult areas and has smoother and flatter edges.

TABLE III
EXPERIMENTAL RESULTS OF DIFFERENT METHODS ON THREE DATASETS

| Method | backbone | LoveDA | | | | Munich | | | | Massachusetts | | | |
|---|---|---|---|---|---|---|---|---|---|---|---|---|---|
| | | IOU(%)/ F1-score(%) / Recall(%)/ Precision(%) | | | | | | | | | | | |
| FCN[19] | Resnet34 | 47.62 | 56.27 | 50.10 | 78.28 | 66.41 | 78.40 | 74.12 | 83.22 | 62.51 | 75.66 | 76.01 | 77.29 |
| Unet[5] | -- | 48.63 | 59.13 | 52.46 | 79.43 | 79.90 | 87.99 | 87.02 | 90.20 | 63.15 | 76.12 | 73.24 | 81.55 |
| PSPNet[3] | Resnet50 | 47.84 | 56.25 | 50.88 | 76.93 | 75.45 | 84.85 | 82.59 | 88.50 | 65.21 | 77.71 | 78.66 | 78.21 |
| RefineNet[4] | Resnet101 | 51.11 | 62.68 | 56.92 | 79.46 | 79.66 | 87.91 | 84.14 | 92.04 | 64.92 | 77.34 | 74.17 | **83.13** |
| DeepLab v3+[2] | Xception101 | 49.80 | 61.55 | 55.89 | 78.25 | 80.84 | 88.63 | 86.35 | 91.98 | 66.68 | 78.88 | 78.91 | 80.35 |
| RoadNet[12] | -- | 49.73 | 61.04 | 54.42 | 79.80 | 80.77 | 88.55 | 85.92 | 92.47 | 63.62 | 76.44 | 72.69 | 82.79 |
| MSMT-RE[10] | -- | 50.50 | 62.91 | 58.83 | 77.32 | 81.51 | 89.06 | 86.59 | 92.49 | 66.31 | 78.51 | 77.01 | 81.82 |
| SRCnet[9] | -- | 52.09 | 63.32 | 58.02 | **80.42** | 82.34 | 89.68 | 88.10 | 91.91 | 67.16 | 78.82 | 78.68 | 80.15 |
| RUnet[8] | -- | 50.15 | 61.86 | 56.31 | 80.37 | 82.33 | 89.66 | 88.01 | 92.32 | 67.04 | 78.93 | 77.30 | 82.45 |
| Baseline (E+D+CGM(base)) | -- | 49.75 | 59.21 | 53.40 | 77.95 | 80.81 | 88.05 | 85.81 | 91.55 | 65.58 | 77.97 | 77.11 | 80.66 |
| Proposed | -- | **54.54** | **66.34** | **62.38** | 80.18 | **84.07** | **90.80** | **89.84** | **92.72** | **68.21** | **80.01** | **80.16** | 81.30 |

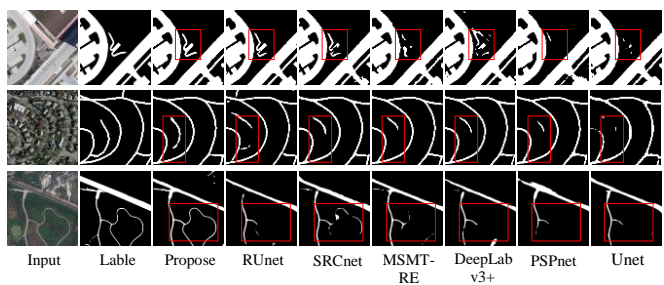

Input　Lable　Propose　RUnet　SRCnet　MSMT-RE　DeepLab v3+　PSPNet　Unet

**Fig 7.** Road detection results from the different methods. The results of the experiments on the Munich Road Dataset, the Massachusetts Road Dataset, and the LoveDA Dataset are shown in the first, second, and third rows, respectively

## IV. CONCLUSION

In this letter, we propose a dual-task network (DTnet) based on feature bridge modules (FBM) and cross-layer graph fusion module (CGM). We use different training tasks in DTnet's two branches to generate various features, increase feature richness and robustness, and then introduce FBM and CGM among two branches features and cross-layer features to optimize their fused feature performance respectively. The experiment results show that the combination of DTnet, FBM, and CGM outperforms the mainstream approaches. In the future, we will further explore the relationship between other more different tasks to achieve a mutually reinforcing effect.